\documentclass{article}
\usepackage{kr}

\usepackage{times}
\usepackage{soul}
\usepackage{url}
\usepackage[hidelinks]{hyperref}
\usepackage[utf8]{inputenc}
\usepackage[small]{caption}
\usepackage{graphicx}
\usepackage{amsmath}
\usepackage{amsthm}
\usepackage{booktabs}
\usepackage{algorithm}
\usepackage{algorithmic}
\usepackage{graphicx} 
\usepackage{amssymb}
\usepackage{multirow}
\usepackage{boldline}
\usepackage{rotating}
\usepackage{adjustbox}

\title{RLGNet: Repeating-Local-Global History Network for Temporal Knowledge Graph Reasoning}

\author{%
Ao Lv$^\dag$\and
Guige Ouyang$^\dag$\and
Yongzhong Huang$^{*}$\and
Yue Chen\and
Haoran Xie\\
($^*$ means to be the corresponding author, $^\dag$ means these authors contributed equally to this work.)\\
\affiliations
School of Computer Science and Information Security, Guilin University Of Electronic Technology, Guilin, China
\emails
22032303175@malis.guet.edu.cn,
\{1045628007,2389483289,592736291\}@qq.com,
21032202040@malis.guet.edu.cn
}

\begin{document}

\maketitle

\begin{abstract}
	Temporal Knowledge Graph (TKG) reasoning involves predicting future events based on historical information. However, due to the unpredictability of future events, this task is highly challenging. To address this issue, we propose a multi-scale hybrid architecture model based on ensemble learning, called RLGNet (\textbf{R}epeating-\textbf{L}ocal-\textbf{G}lobal History Network). Inspired by the application of multi-scale information in other fields, we introduce the concept of multi-scale information into TKG reasoning. Specifically, RLGNet captures and integrates different levels of historical information by combining modules that process information at various scales. The model comprises three modules: the Repeating History Module focuses on identifying repetitive patterns and trends in historical data, the Local History Module captures short-term changes and details, and the Global History Module provides a macro perspective on long-term changes. Additionally, to address the limitations of previous single-architecture models in generalizing across single-step and multi-step reasoning tasks, we adopted architectures based on Recurrent Neural Networks (RNN) and Multi-Layer Perceptrons (MLP) for the Local and Global History Modules, respectively. This hybrid architecture design enables the model to complement both multi-step and single-step reasoning capabilities. Finally, to address the issue of noise in TKGs, we adopt an ensemble learning strategy, combining the predictions of the three modules to reduce the impact of noise on the final prediction results. In the evaluation on six benchmark datasets, our approach generally outperforms existing TKG reasoning models in multi-step and single-step reasoning tasks.
\end{abstract}

\begin{figure}[t]
	\centering
	\includegraphics[width=\linewidth]{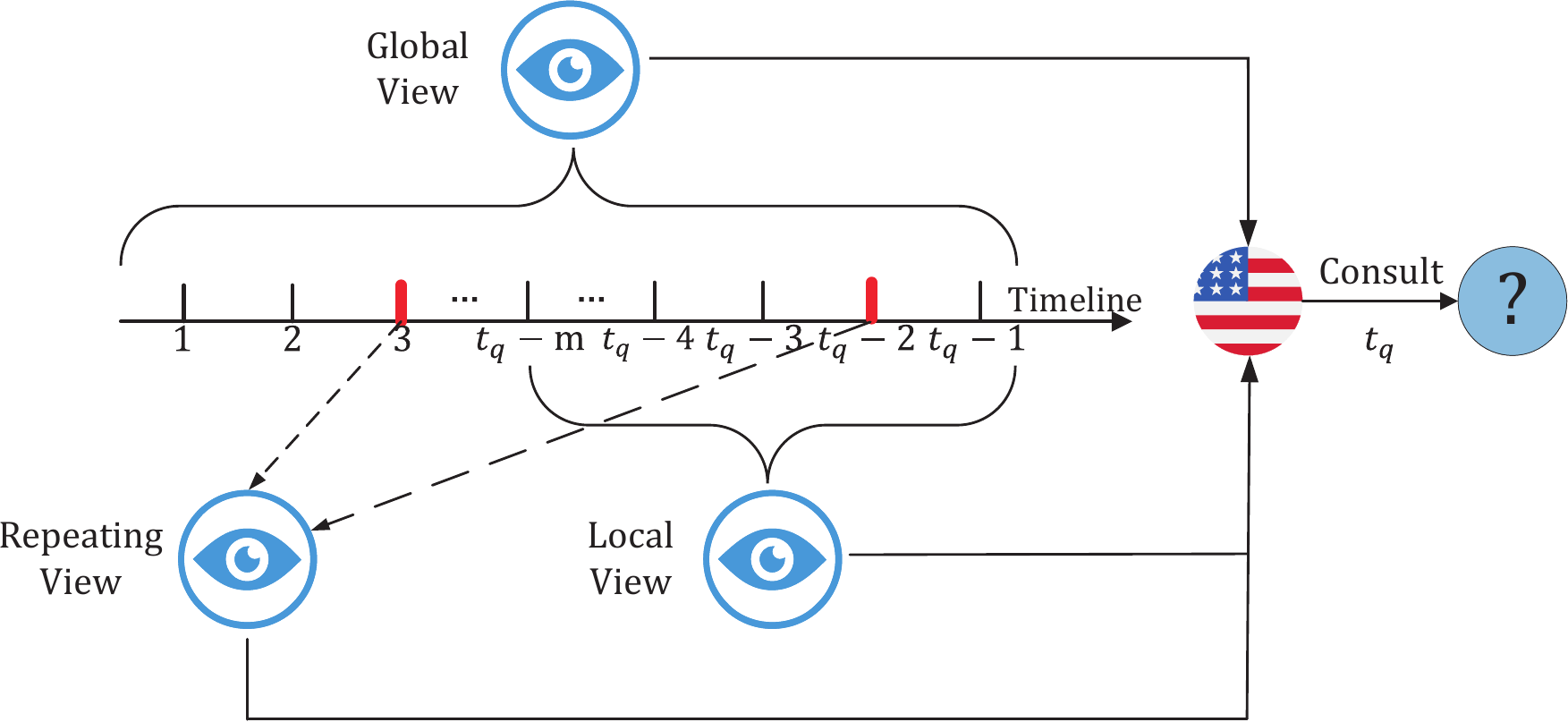}
	\caption{The figure illustrates the repeating, local, and global perspectives. The red timestamp indicates that an fact identical to the queried fact occurred at that specific moment in time.}
	\label{fig:1}
\end{figure}

\section{Introduction}
	A Temporal Knowledge Graph (TKG) is a structured yet highly complex knowledge system. In TKG, each fact is represented by a quadruple consisting of subject, relation, object, and timestamp. This paper primarily discusses the problem of extrapolation in TKG. TKG allows us to understand the relationships between entities and capture changes in these relationships over time. The goal of TKG extrapolation is to predict new facts at future time steps based on known facts. Extrapolation can be further divided into single-step and multi-step types. Single-step reasoning involves predicting facts that will occur at the current time step (time $t$) based on known past facts. Multi-step reasoning requires sequentially inferring facts at multiple future time steps, such as $t$, $t+1$, $t+2$, and so on. However, the inclusion of unknown factors in future events poses significant challenges for TKG extrapolation \cite{Re-net}.

	Recent studies indicate that some facts in TKG exhibit traceable patterns, hence simulating the evolution of these facts can aid in predicting them. For instance, models based on RNN architecture are widely used \cite{RE-GCN}. Additionally, there are models based on MLP \cite{CyGNet}, where time is often vectorized and input as a feature to capture the temporal characteristics of facts. To achieve more accurate predictions, some models cleverly design methods to fuse local and global historical information, allowing the model to have both global and local perspectives. The global perspective helps the model better generalize the entire dataset's characteristics, while the local perspective assists in extracting recent fact information. Furthermore, some studies identify closely related entities as candidate answers, using them to constrain the prediction range and improve accuracy.
	
	Beyond these designs, we further observe that a portion of facts in TKG are closely related to the past, as evidenced by the significant proportion of facts that repeat multiple times in TKG. Thus, when predicting the future, it is reasonable to assume that these facts will reoccur. Conversely, some facts in TKG only occur at specific times, contributing to substantial noise in TKG datasets. This noise causes models to fit these time-specific facts during training, impairing their ability to predict future events. In TKG multi-step reasoning tasks, the absence of facts over a period can cause RNN-based models to evolve incorrect facts. Over time, the influence of past facts diminishes, and MLP-based models tend to assign similar weights to facts at different times, affecting their performance in single-step reasoning. Just as RNN and MLP models each have their strengths in single-step and multi-step reasoning tasks, they also have advantages in capturing historical information at different scales. MLP-based models are well-suited for capturing patterns and relationships across the entire dataset because they process each input independently, without temporal bias, allowing them to generalize long-term trends and patterns. On the other hand, RNN-based models are ideal for capturing local historical information, as they excel at handling short sequences of data and can effectively capture short-term changes in the most recent time steps.
	
	Based on these observations, we designed a model called the \textbf{R}epeating-\textbf{L}ocal-\textbf{G}lobal History Network (RLGNet). Specifically, RLGNet draws inspiration from ensemble learning by designing three relatively independent modules: Global History Module, Local History Module, and Repeating History Module. As shown in Figure \ref{fig:1}, these three modules capture information at different scales. During training, RLGNet first learns the Local and Global History Modules in parallel, while the Repeating History Module learns serially on the results of the Global and Local History Modules to further reinforce the scores of repeating facts. This ensemble learning strategy can reduce the sensitivity of individual models to noisy data in TKG, better capture the true patterns in the data, and reduce the risk of overfitting. Due to the advantages of RNN-based and MLP-based models in single-step and multi-step reasoning tasks, as well as their effectiveness in handling historical information at different scales, RLGNet adopts an RNN-based structure for the Local History Module and an MLP-based structure for the Global History Module.  This hybrid design fully leverages the complementary strengths of both approaches. Finally, to further constrain the answer range, we integrate candidate entities' information into the three learning processes. Our multi-scale hybrid architecture design enables RLGNet to achieve state-of-the-art performance in most cases for both multi-step and single-step reasoning tasks. Our contributions are summarized as follows:
	\begin{enumerate}
	\item Unlike previous work, we designed an ensemble learning-based model that processes different historical information from a multi-scale perspective. Specifically, we use Repeating, Local, and Global History Modules to capture their corresponding historical information, and these modules collaborate to enhance the overall performance of the model.
	\item We use RNN and MLP architectures in the Local History Module and Global History Module, respectively, and integrate their scores to achieve complementary capabilities between the two modules. This hybrid architecture model addresses the limitation of previous works, where a single architecture could only achieve good results in either multi-step or single-step tasks, improving the model's performance in multi-task scenarios.
	\item We proposed a multi-scale hybrid architecture model based on an ensemble learning strategy, which demonstrated impressive performance across six public TKG datasets.
	\end{enumerate}

\section{Related Work}

\textbf{Static KG Reasoning.} Recent years have seen great interest and research in static Knowledge Graph (KG) reasoning models.\cite{ComlEX,TTransE} These models include distance-based models, such as TransE\cite{TransE} and TransH\cite{TransH}, which determine the likelihood of facts by measuring the distances between entities. Another category is semantic-matching-based models, like DistMult\cite{DistMult} and RESCAL\cite{RESCAL}. There are also models based on Convolutional Neural Networks (CNNs), such as ConvE\cite{ConvE} and Conv-TransE\cite{Conv-TransE}, which represent entities and relations using matrices processed by convolutional kernels. Similarly, Graph Convolutional Network (GCN) models, including R-GCN\cite{R-GCN} and VR-GCN\cite{VR-GCN}, stand out for their capacity to integrate graph structures with node features.
However, these models focus on static KG, and their predictive capabilities for future events are limited.

\textbf{Temporal KG Reasoning.} TKG reasoning has two settings: extrapolation and interpolation. The interpolation setting aims at predicting missing historical facts rather than future events.
In contrast, this paper focuses on reasoning within the extrapolation setting to forecast future facts using historical data.
Know-Evolve\cite{Know-Evolve} and DyREP\cite{DyRep} model the occurrence of facts in TKG using a temporal point process. Glean\cite{Glean} enriches factual features by utilizing unstructured text information. 
CyGNet\cite{CyGNet} captures and understands historical trends and patterns through a replicating generative mechanism. CENET\cite{CENET} distinguishes events as historical and non-historical, using this classification for contrastive learning. Some models, such as RE-GCN\cite{RE-GCN}, utilize Graph Convolutional Networks (GCNs) to simulate the evolutionary process of knowledge graphs, thereby capturing and learning the dynamic properties of entities and relations.
TiRGN\cite{TiRGN} uses a local-global historical approach for reasoning. CEN\cite{CEN} and CluSTeR\cite{CluSTeR} utilize reinforcement learning to capture and learn the connections between entities and relationships in the knowledge graph. TLogic\cite{TLogic} utilizes temporal logic rules to constrain the predicted paths for queries. TANGO\cite{TANGO} utilizes neural ordinary differential equations to model the structural information of each entity. However, these models often focus on partial aspects of historical information without combining query data with historical details. Consequently, they can't accurately capture recurring, local, and global facts.

\section{Preliminaries}

TKGs represent events temporally as snapshots. Let $\mathcal{G}_{raw}$ be the sequence of snapshot graphs $\mathcal{G}_{raw}=\{\mathcal{G}_{raw}^{1},\dots,\mathcal{G}_{raw}^{\vert \mathcal{T} \vert} \}$. Each fact in the snapshot is a quadruple $(s,r,o,t)$, where $s,o \in \mathcal{E},r \in \mathcal{R},t \in \mathcal{T}$. It represents a relationship $r$ between subject entity $s$ and object entity $o$ at time $t$. $\mathcal{E}$, $\mathcal{R}$, and $\mathcal{T}$ represent the sets of entities, relations, and timestamps, respectively.
In this paper, our reasoning task is to predict missing entities, specifically the subject entity $s$ and the object entity $o$. To simplify the problem, we convert the task of predicting the object entity $o$ into predicting the subject entity $s$, represented as $(o,r^{-1},?,t)$. Therefore, for each quadruple $(s,r,o,t)$, we add a reverse relation quadruple $(o,r^{-1},s,t)$ to the dataset.

Given the query tuple $q=(s_q,r_q,?,t_q)$, we define the candidate entity set at time $t$ as $\mathcal{C}_q^t=\{o\vert(s_q,r_q,o,t)\in \mathcal{G}^t_{raw}\}$. The candidate entity set over a time range $t_1$ to $t_2$ is defined as $\mathcal{C}_q^{t_1:t_2}$, a union of all candidate sets over the range:
\begin{equation}\label{eq1}
	\mathcal{C}_q^{t_1:t_2}=\bigcup_{i=t_1}^{t_2}\mathcal{C}_q^i
\end{equation}

Since not all candidate entities greatly affect the prediction, we only keep the top $top_k$ most frequent entities, denoted as $\mathcal{C}_{q,top_k}^{1:t_q-1}$. To represent repeating facts, we need to select those facts that have occurred more than once in history and define repeating facts that occurred at time $t$ as $\mathcal{G}_{rep}^{t}$.
$\mathcal{G}_{rep}^{t}$ can be formalized as:

\begin{equation}\label{eq2}
	\mathcal{G}_{rep}^{t}=\{q'\vert q'=(s,r,o,t) \in \mathcal{G}^{t}_{raw}, o \in \mathcal{C}_{q',top_k}^{1:t-1}\}
\end{equation}

Similar to $\mathcal{G}_{raw}$, we obtain $\mathcal{G}_{rep}=\{\mathcal{G}_{rep}^{1},\dots,\mathcal{G}_{rep}^{\vert \mathcal{T} \vert}\}$.

\begin{figure*}[t]
	\centering
	\includegraphics[width=\linewidth]{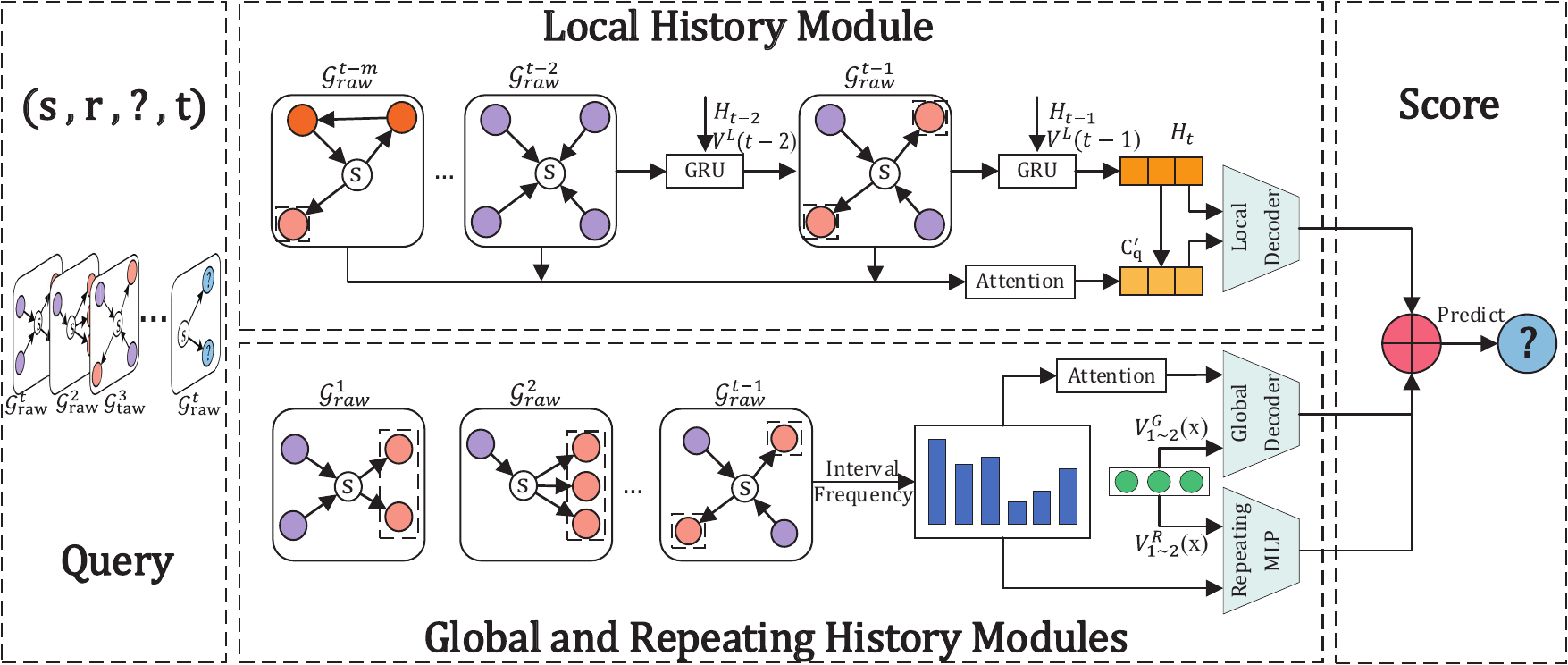}
	\caption{The upper half represents the Local History Module, which learns local historical information through KG sequences of adjacent timestamps. The lower half comprises the Repeating and Global History Modules, learning repetitive and global historical information respectively by statistically querying candidate entities. The entities within the dashed box are candidate entities. The left side weights and sums the scores of the three modules to obtain the final prediction score.}
	\label{fig:2}
\end{figure*}

\section{Model Overview}

The overall framework of the model is illustrated in Figure \ref{fig:2}. This model can be divided into three core submodules: the Local History Module, the Global History Module, and the Repeating History Module. The Local and Global History Modules are used to capture historical information at local and global scales, respectively, with the Local History Module based on an RNN architecture and the Global History Module based on an MLP architecture. The Repeating History Module is used to further enhance the scores of repeating facts. To handle the frequency of time and entity occurrences, we use a Numerical Embedding function to convert them into embeddings. To further constrain the answer range, we apply the candidate entity design to all three modules. The model is designed based on the concept of ensemble learning, so the repeating, global, and local modules are trained independently, and their prediction scores are combined through weighted summation to produce the final result.

\subsection{Numerical Embedding}
The Numerical Embedding module is used to convert time and entity occurrence frequency into embeddings. The cosine function can capture periodic patterns in time and frequency, while the tanh function can provide smoother transitions for phenomena with rapidly changing frequencies or sudden changes in time.
\begin{equation}\label{eq4}
	\left\{\begin{aligned}
		\textbf{v}_1 (x)&=\cos(W_1^Nx+b_1^N)\\
		\textbf{v}_2 (x)&=\tanh(W_2^Nx+b_2^N)
	\end{aligned}\right.
\end{equation}
Then we concatenate the two embeddings to obtain the final embedding.
\begin{equation}\label{eq5}
	\textbf{V}(t)= [\textbf{v}_1(x);\textbf{v}_2(x)]
\end{equation}
Here, $W_1^N,W_2^N,b_1^N,b_2^N$ are learnable parameters and the embeddings $\textbf{v}_1(x)$ and $\textbf{v}_2(x)$ have the same dimensionality. The notation [;] denotes the embedding concatenation operation, and $\textbf{V}(x)$ represents the Numerical Embedding. Combining the embeddings generated by these two modules with different characteristics allows for the integration of both periodic and nonlinear information, providing a more comprehensive representation.

\subsection{Local History Module}
This module captures local facts by focusing on adjacent history. For each query $q=(s_q,r_q,?,t_q)$, the module pays attention to the $m$ timestamp subgraph $\{\mathcal{G}^{t_q-m}_{raw},\dots,\mathcal{G}^{t_q-1}_{raw}\}$ related to this query to obtain the structural features of the subgraph. Here, $m$ is a hyperparameter. To achieve this, we employ a GCN to aggregate individual subgraphs, use Gated Recurrent Units (GRU) \cite{GRU} to learn the features of subgraph evolution, and adopt an attention mechanism to integrate information from candidate entities.
Firstly, we use a GCN with $\omega$ layers to get the entity embeddings at the current moment:
\begin{equation}\label{eq6}
	h_o^{t,l+1}=\sum\limits_{(s,r,o,t) \in \mathcal{G}^{t}_{raw}}\frac{1}{\vert N_o^t\vert}W_1^l \phi(h_s^{t,l},r^L)+W_2^l h_o^l
\end{equation}
\begin{equation}\label{eq6.1}
	\phi(h_s^{t,l},r^L)=[h_s^{t,l};r^L;h_s^{t,l}+r^L;h_s^{t,l}\cdot r^L ]
\end{equation}
Here, the symbol [$\cdot$] represents the Hadamard product. $N_o^t$ represents the neighbors of node $o$ at time $t$. The embedded of node $o$ and $s$ at the $l$-th layer is denoted as $h_o^{t,l}$ and $h_s^{t,l}$. The aggregation and self-loop parameters of the $l$-th layer are denoted as $W_1^l$ and $W_2^l$. $r^L$ is the initial embedding of the relation in the Local History Module. When $\omega =0$,$h_o^{t,0}=\sum_{i \in N_o^t}\frac{1}{\vert N_o^t\vert} h_i^t$. If $t =t_q-m$ then $h^{t_q-m}_{o}$ represents the initial embedding of the entity $o$ in the Local History Module. Entity embeddings are combined with their temporal embeddings and then predicted for the next moment with a GRU:
\begin{equation}\label{eq7}
	H_{t+1}=GRU([H_t;\textbf{V}^L(t_q-t)],H^{'}_t)
\end{equation}
Here, $\textbf{V}^L$ is Numerical Embeddings. $H^{'}_t$ is the output of the last layer of the GCN for all entities at time $t$, and $H_{t}$ is the embedding representation of all entities at time $t$. The attention mechanism aggregates candidate entities over a history of length $m$. Node aggregation for query $q$ is:
\begin{equation}\label{eq8}
	C_q^t=\sum\limits_{i \in \mathcal{C}_q^t}\frac{1}{\vert \mathcal{C}_q^t \vert} h_i^{t}
\end{equation}
Here, $C_q^t$ represents query $q$'s candidate entity embedding at time $t$. $h_i^{t}$ represents the embedding of candidate entity $i$ in $H_t$. Next, the module uses a 2-layer MLP to calculate the attention weights $a_q^t$ for the query at time $t$:
\begin{equation}\label{eq9}
	a_q^t=MLP([h_q^{t};r_q^L;\textbf{V}^L(t_q-t);C_q^t])
\end{equation}
Here, $r_q^L$ is the initial embedding of the query relation in the Local History Module. $h_q^{t}$ represents the embedding of the query entity in $H_t$. After calculating attention weights, these are used to aggregate candidate entities for query q:
\begin{equation}\label{eq10}
	C_q^{'}=\sum_{i=t_q-m}^{t_q-1}\frac{\exp{(a_q^i)}\cdot C_q^i }{\sum_{j=t_q-m}^{t_q} \exp{(a_q^j)} }
\end{equation}
This results in the candidate entity embedding $C_q^{'}$ for query $q$. The Local History Module uses ConvTransE as the scoring function. To better adapt to the task, we additionally concatenate query-related information embeddings to the original entity and relation embeddings in ConvTransE. These concatenated embeddings are then processed through a one-dimensional convolution layer followed by a fully connected layer to obtain the scores:
\begin{equation}\label{eq11}
	score_{loc}^q=H_{t_q} \cdot \phi_{L}(h_q^{t_q},r_q^L,C_q^{'})
\end{equation}
Here, $\phi_{L}$ represents the score decoder based on ConvTransE in the Local History Module.

\subsection{Global History Module}

The Global History Module is designed to consider and extract information from a global perspective in order to capture global facts. For each query, we calculate the occurrence frequency of candidate entities and the last time they appeared. The module then uses an attention mechanism to integrate this information about the candidate entities.
\begin{equation}\label{eq12}
	\mathcal{A}_{1} (q,i)=(W_1^G[h_q^G;r_q^G])^T (W_2^G [\textbf{V}_1^G(t_q-t_q^i );h_i^G])
\end{equation}
The attention scores for frequency-related candidates are calculated similarly:
\begin{equation}\label{eq13}
	\mathcal{A}_{2} (q,i)=(W_3^G[h_q^G;r_q^G])^T (W_4^G [\textbf{V}_2^G(cnt^i_q);h_i^G])
\end{equation}
where $h_q^G$, $r_q^G$, and $h_i^G$ refer to the initial embeddings of the query entity, relation, and candidate entity $i$ in the Global History Module. $\textbf{V}_1^G$ and $\textbf{V}_2^G$ are Numerical Embeddings. $W_1^G$, $W_2^G$, $W_3^G$, and $W_4^G$ are learnable parameters. $t_q^i$ and $cnt_q^i$ represent the last occurrence and frequency of candidate entity $i$ for query $q$. Additionally, taking computational resources into consideration, we only selected the $top_k^{all}$ candidate entities. Following this, we normalize their respective scores and multiply by the candidate entity embeddings to obtain the embedding $C_q^{gap}$ for time-relevant:
\begin{equation}\label{eq14}
	C_q^{gap}=\sum_{i \in \mathcal{C}_{q,top_k^{all}}^{1:t_q-1}}\frac{\exp(\mathcal{A}_{1} (q,i))\cdot h_i^G}{\sum\limits_{j \in \mathcal{C}_{q,top_k^{all}}^{1:t_q-1}}\exp (\mathcal{A}_{1} (q,j))}
\end{equation}
We normalize $\mathcal{A}_{2}$ and multiply it by the entity embeddings, then sum the results to obtain the frequency-related candidate embedding $C_q^{cnt}$, just as we do with $C_q^{gap}$. Similar to Equation \ref{eq11}, the score for the Global History Module is:
\begin{equation}\label{eq16}
	score_{Glo}^q=H_G \cdot \phi_{G} (h_q^G,r_q^G,\textbf{V}_3^G(t_q),C_q^{gap},C_q^{cnt} )
\end{equation}
where $H_G$ represents the initial embeddings of all entities in the Global History Module. $\textbf{V}_3^G$ is Numerical Embeddings. $\phi_{G}$ represents the score decoder based on ConvTransE in the Global History Module.

\subsection{Repeating History Module}

The Repeating History Module is designed to improve prediction by increasing the weight of historical events. We first filter and retain the repeated facts, and refer to the collection of these repeated facts as $\mathcal{G}_{rep}$. Then, only the candidate entities of facts that rank in the top $top_k$ in terms of frequency are retained.

The Repeating History Module's score function is a three-layer MLP. The candidate entities' scores are calculated using the MLP as follows:
\begin{equation}\label{eq18}
	score_{Rep}^{q,i}=MLP([h_q^R;r_q^R;\textbf{V}_1^R(t_q);h_i^R;\textbf{V}_2^R(cnt_q^i )])
\end{equation}
where $h_q^R$,$h_i^R$ and $r_q^R$ represent the initial embeddings of the query entity, candidate entity, and relation in the Repeating History Module, respectively. $\textbf{V}_1^R$ and $\textbf{V}_2^R$ are Numerical Embeddings. It should be noted that when candidate entity $i\notin \mathcal{C}_{q,top_k}^{1:t_q-1}$, the $score_{Rep}^{q,i}$ is 0.

\subsection{Loss Function}

Since the model employs an ensemble learning strategy, each module is trained independently and uses the cross-entropy loss function to calculate the loss value:
\begin{equation}\label{eq19}
	\mathcal{L}(score,\mathcal{G})=\sum_{\mathcal{G}^t \in \mathcal{G}} \sum_{(s,r,o,t) \in \mathcal{G}^t} y_t\log P(o|s,r,t)
\end{equation}
The score represents the predicted scores of each module, and $\mathcal{G}$ represents the sequence of snapshot graphs. Here, $P(o|s,r,t)=\text{softmax}(score)$ represents the predicted probability of entities, and $y_t \in \vert\mathcal{E}\vert$ is the label vector, where an element is 1 if the fact occurs, or 0 otherwise. Therefore, the loss functions for the three modules are $\mathcal{L}(score_{Loc}, \mathcal{G}_{raw})$, $\mathcal{L}(score_{Glo}, \mathcal{G}_{raw})$, and $\mathcal{L}(score_{Rep}, \mathcal{G}_{rep})$, respectively.

\subsection{Final Score}	

The final model score combines scores from the Repeating, Local, and Global History Modules:
\begin{equation}\label{eq20}
	\begin{split}
		score_{Fin}^q =& \ \alpha \cdot score_{Loc}^q+(1-\alpha)\cdot score_{Glo}^q\\ &+score_{Rep}^q
	\end{split}
\end{equation}
Here, $\alpha \in [0,1]$ is a hyperparameter.

\begin{table*}[t]
	\centering
	\begin{tabular}{c c c c c c c c c c c c c c}
		\toprule
		\multicolumn{2}{c}{\multirow{2}{*}{Model}} & \multicolumn{4}{c}{ICEWS18} & \multicolumn{4}{c}{ICEWS14} & \multicolumn{4}{c}{ICEWS05-15} \\ \cmidrule(lr){3-6}\cmidrule(lr){7-10}\cmidrule(lr){11-14}
		& & MRR & H@1 & H@3 & H@10 & MRR & H@1 & H@3 & H@10 & MRR & H@1 & H@3 & H@10 \\ \hline
		\multirow{6}{*}{\begin{turn}{90}Single-Step\end{turn}}&xERTE & 29.31 & 21.03 & 33.51 & 46.48 & 40.79 & 32.70 & 45.67 & 57.30 & 46.62 & 37.84 & 52.31 & 63.92 \\ 
		&RE-GCN & 32.62 & 22.39 & 36.79 & 52.68 & 42.00 & 31.63 & 47.20 & 61.65 & 48.03 & 37.33 & 53.90 & 68.51 \\ 
		&TITer & 29.98 & 22.05 & 33.46 & 44.83 & 41.73 & 32.74 & 46.46 & 58.44 & 47.60 & 38.29 & 52.74 & 64.86 \\ 
		&TiRGN & 33.66 & 23.19 & 37.99 & 54.22 & 44.04 & 33.83 & 48.95 & 63.84 & 50.04 & 39.25 & \textbf{56.13} & \textbf{70.71} \\ 
		&CEN & 31.50 & 21.70 & 35.44 & 50.59 & 42.20 & 32.08 & 47.46 & 61.31 & 46.84 & 36.38 & 52.45 & 67.01 \\ 
		&RETIA & 32.43 & 22.23 & 36.48 & 52.94 & 42.76 & 32.28 & 47.77 & 62.75 & 47.26 & 36.64 & 52.90 & 67.76 \\ \hline
		&RLGNet & \textbf{34.96} & \textbf{24.68} & \textbf{39.22} & \textbf{55.09} & \textbf{46.15} & \textbf{36.16} & \textbf{51.17} & \textbf{65.12} & \textbf{50.56} & \textbf{40.34} & 56.05 & 70.18 \\ \hline
		\multirow{4}{*}{\begin{turn}{90}Multi-Step\end{turn}}&CyGNet & 26.07 & 16.76 & 29.54 & 44.43 & 34.80 & 25.34 & 39.05 & 53.09 & 38.17 & 27.93 & 43.01 & 57.89 \\ 
		&RE-GCN & 28.44 & 19.03 & 31.96 & 46.86 & 37.68 & 28.00 & 41.81 & 56.87 & 38.74 & 28.50 & 43.60 & 58.52 \\ 
		&TiRGN & 28.85 & 19.18 & 32.58 & 47.78 & 38.37 & 28.80 & 42.50 & 56.94 & 39.97 & 29.44 & 44.76 & 60.92 \\ 
		&CENET & 27.40 & 18.91 & 30.26 & 44.36 & 35.62 & 27.10 & 38.81 & 52.31 & 39.92 & 30.21 & 44.14 & 59.09 \\ \hline
		&RLGNet & \textbf{29.90} & \textbf{20.18} & \textbf{33.64} & \textbf{49.08} & \textbf{39.06} & \textbf{29.34} & \textbf{42.03} & \textbf{58.12} & \textbf{40.83} & \textbf{30.06} & \textbf{45.91} & \textbf{61.93} \\ 
		\bottomrule
	\end{tabular}
	\caption{Performance (in percentage) on ICEWS18, ICEWS14, and ICEWS05-15.}
	\label{tab:2}
\end{table*}

\begin{table*}[t]
	\centering
	\begin{tabular}{c c c c c c c c c c c c c c}
		\toprule
		\multicolumn{2}{c}{\multirow{2}{*}{Model}} & \multicolumn{4}{c}{WIKI} & \multicolumn{4}{c}{YAGO} & \multicolumn{4}{c}{GDELT} \\  
		\cmidrule(lr){3-6}\cmidrule(lr){7-10}\cmidrule(lr){11-14}
		& & MRR & H@1 & H@3 & H@10 & MRR & H@1 & H@3 & H@10 & MRR & H@1 & H@3 & H@10 \\ \hline
		\multirow{6}{*}{\begin{turn}{90}Single-Step\end{turn}} &xERTE & 73.60 & 69.05 & 78.03 & 79.73 & 84.19 & 80.09 & 88.02 & 89.78 & 19.45 & 11.92 & 20.84 & 34.18 \\
		&RE-GCN & 78.53 & 74.50 & 81.59 & 84.70 & 82.30 & 78.83 & 84.27 & 88.58 & 19.69 & 12.46 & 20.93 & 33.81 \\ 
		&TITer & 73.91 & 71.70 & 75.41 & 76.96 & 87.47 & 84.89 & 89.96 & 90.27 & 18.19 & 11.52 & 19.20 & 31.00 \\ 
		&TiRGN & 81.65 & 77.77 & 85.12 & 87.08 & 87.95 & 84.34 & 91.37 & 92.92 & 21.67 & 13.63 & 23.27 & 37.60 \\ 
		&CEN & 78.93 & 75.05 & 81.90 & 84.90 & 83.49 & 79.77 & 85.85 & 89.92 & 20.39 & 12.96 & 21.77 & 34.97 \\ 
		&RETIA & 78.59 & 74.85 & 81.39 & 84.58 & 81.04 & 77.00 & 83.31 & 88.62 & 20.12 & 12.76 & 21.45 & 34.49 \\ \hline
		&RLGNet & \textbf{82.43} & \textbf{78.86} & \textbf{85.65} & \textbf{87.17} & \textbf{89.69} & \textbf{87.05} & \textbf{92.15} & \textbf{93.00} & \textbf{25.09} & \textbf{16.95} & \textbf{27.42} & \textbf{40.87}\\ \hline 
		\multirow{4}{*}{\begin{turn}{90}Multi-Step\end{turn}}&CyGNet & 58.44 & 53.03 & 62.24 & 67.46 & 68.60 & 60.97 & 73.58 & 83.16 & 19.11 & 11.90 & 20.31 & 33.12 \\ 
		&RE-GCN & 62.05 & 58.95 & 63.89 & 67.39 & 70.05 & 65.76 & 72.70 & 77.16 & 19.62 & 12.47 & 20.86 & 33.48 \\ 
		&TiRGN & 64.04 & 60.72 & 66.52 & 68.96 & 78.51 & 74.01 & 82.74 & 84.76 & 19.87 & 12.46 & 21.21 & 34.25 \\ 
		&CENET & 57.52 & 51.99 & 61.93 & 66.29 & 69.90 & 64.01 & 73.04 & 82.65 & - & - & - & - \\ \hline 
		&RLGNet & \textbf{64.34} & \textbf{61.03} & \textbf{66.71} & \textbf{69.51} & \textbf{80.17} & \textbf{76.52} & \textbf{83.57} & \textbf{84.96} & \textbf{20.81} & \textbf{13.34} & \textbf{22.32} & \textbf{35.38} \\ 
		\bottomrule
	\end{tabular}
	\caption{Performance (in percentage) on WIKI, YAGO, and GDELT.}
	\label{tab:3}
\end{table*}

\section{Experiments}

\begin{table*}[t]
	\centering
	\begin{tabular}{ c c c c c c c }
		\toprule
		Dataset & $\vert\mathcal{E}\vert$ & $\vert\mathcal{R}\vert$ & Train & Valid & Test & Time gap \\ \hline
		ICEWS18 & 23,033 & 256 & 373,018 & 45,995 & 49,545 & 24 hours \\ 
		ICEWS14 & 7,128 & 230 & 63,685 & 13,823 & 13,222 & 24 hours \\ 
		ICEWS05-15 & 10,488 & 251 & 368,868 & 46,302 & 46,159 & 24 hours \\
		WIKI & 12,554 & 24 & 539,286 & 67,538 & 63,110 & 1 year \\ 
		YAGO & 10,623 & 10 & 161,540 & 19,523 & 20,026 & 1 year \\ 
		GDELT & 7,691 & 240 & 1,734,399 & 238,765 & 305,241 & 15 mins \\ 
		\bottomrule
	\end{tabular}
	\caption{Statistics of the datasets.}
	\label{tab:1}
\end{table*}

\subsection{Setup}
\subsubsection{Datasets} 
We use six TKG datasets to evaluate the model's effectiveness in the entity prediction task, including ICEWS14 \cite{Alberto2018}, ICEWS18 \cite{ICEWS18}, and ICEWS05-15 \cite{Alberto2018} from the Integrated Crisis Early Warning System (ICEWS), and the event-driven GDELT \cite{GDELT} dataset. Public datasets WIKI \cite{WIKI} and YAGO \cite{YAGO} are also included. All datasets are time-partitioned into Training (80\%), Validation (10\%), and Test (10\%). More details on the datasets are provided in Table \ref{tab:1}.

\subsubsection{Evaluation Metrics} 
To assess TKG reasoning performance, we use Mean Reciprocal Rank (MRR) and Hits@k metrics. MRR calculates the average inverse rankings for actual entities across all queries, while Hits@k denotes the proportion of real entities appearing within the top k rankings. Therefore, higher values of MRR and Hits@k indicate better model performance. Previous studies have indicated that traditional filtering settings are flawed \cite{TLDR}. Thus, we report the experimental results after using the time-aware filtering settings.

\subsubsection{Baselines} 
RLGNet is compared with six baseline models on the single-step reasoning task, including xERTE \cite{xERTE}, RE-GCN \cite{RE-GCN}, TITer \cite{TITer}, TiRGN \cite{TiRGN}, CEN \cite{CEN}, and RETIA \cite{RETIA}. Since some models are not designed for the multi-step reasoning task, we have selected a few models to report their performance on the multi-step reasoning task, including CyGNet \cite{CyGNet}, RE-GCN \cite{RE-GCN}, TiRGN \cite{TiRGN}, and CENET \cite{CENET}.

\subsubsection{Implementation Details} 
For all datasets, we set the embedding dimension to 200, and the dimension of $\textbf{V}_1^L$ in the Local History Module to 48. The $top_k$ and $top_k^{all}$ are set to 20 and 200, respectively. The number of GCN layers $\omega$ is set to 1. For ICEWS18, ICEWS14, ICEWS05-15, GDELT, WIKI, and YAGO, the adjacent history length $m$ is set to 10, 10, 15, 10, 1, and 1, respectively. The hyperparameter $\alpha$ is set to 0.8 in ICEWS, 0.9 in YAGO and WIKI, and 0.1 in GDELT. Additionally, a static graph constraint similar to that of RE-GCN is also added to ICEWS. Adam is used for parameter learning, with a learning rate set to 0.001. In the Local History Module, we employ StepLR to adjust the learning rate. In this setup, the learning rate decay is set to 0.8. For the YAGO and WIKI datasets, the step-size value is set to 2, while for other datasets, the step-size is 10. Please note that each module is trained independently, which results in them not sharing weights.

\subsection{Results}

The reasoning results of the entity prediction task are shown in Tables \ref{tab:2} and \ref{tab:3}, where RLGNet outperforms other baselines in most cases.On six benchmark datasets, RLGNet has demonstrated superior performance in both multi-step and single-step reasoning tasks in most cases, particularly on the GDELT and ICEWS14 datasets, where its MRR scores have increased by 3.42\% and 2.11\%, respectively. This improvement is primarily due to RLGNet's hybrid architecture and ensemble learning design. Each independent module considers historical information from different scales and constrains the answer range using candidate entities. The hybrid architecture design enhances the model's performance in multi-task scenarios, ultimately leading to more accurate predictions. Compared to various single architecture models such as RE-GCN and TiRGCN, RLGNet exhibits stronger generalization capabilities in both multi-step and single-step reasoning. Moreover, compared to models that do not consider multiple scales, such as CEN and CyGNet, RLGNet shows superior performance.

However, it is worth noting that in the single-step reasoning task on the ICEWS05-15 dataset, RLGNet's H@3 and H@10 metrics are 0.08\% and 0.53\% lower than those of TiRGCN, respectively, but its H@1 metric is 1.09\% higher than that of TiRGCN. We believe this phenomenon is because, in TKG extrapolation tasks, methods that consider candidate entities typically achieve higher H@1 results compared to those that ignore candidate entities. However, these methods may perform slightly worse on other metrics such as H@3 and H@10. This is because introducing candidate entities constrains the answer range, but candidate entities are sampled based on past facts, making the model more inclined towards past patterns. As a result, the model may overlook new emerging patterns or changes when predicting the future. Therefore, although candidate entities can improve the accuracy of the H@1 metric, they may lack diversity and breadth.

\begin{table*}[t]
	\centering
	\begin{tabular}{ c c c c c c c c c c c c c}
		\toprule
		\multirow{2}{*}{Model} & \multicolumn{2}{c}{ICEWS18} & \multicolumn{2}{c}{ICEWS14} & \multicolumn{2}{c}{ICEWS05-15} & \multicolumn{2}{c}{WIKI} & \multicolumn{2}{c}{YAGO} & \multicolumn{2}{c}{GDELT} \\ \cmidrule(lr){2-3}\cmidrule(lr){4-5}\cmidrule(lr){6-7}\cmidrule(lr){8-9}\cmidrule(lr){10-11}\cmidrule(lr){12-13}
		& Single & Multi & Single & Multi & Single & Multi & Single & Multi & Single & Multi & Single & Multi \\ \hline
		\textbf{Glo} & 32.54 & 29.10 & 42.86 & 38.51 & 45.41 & 39.05 & 59.46 & 52.94 & 71.85 & 60.92 & 24.72 & 20.63 \\ 
		\textbf{Loc} & 34.08 & 28.19 & 44.68 & 37.09 & 50.04 & 38.88 & 81.72 & 63.86 & 87.22 & 77.84 & 21.57 & 19.42 \\ 
		\textbf{Glo} + \textbf{Loc} & 34.52 & 29.13 & 45.56 & 38.42 & 50.23 & 39.97 & 81.75 & 63.88 & 87.38 & 78.39 & 24.73 & 20.72 \\ 
		\textbf{Glo} + \textbf{Rep} & 33.19 & 29.56 & 43.74 & 38.64 & 46.57 & 40.14 & 66.75 & 56.98 & 78.85 & 69.29 & \textbf{25.11} & 20.74 \\ 
		\textbf{Loc} + \textbf{Rep} & 34.93 & 29.39 & 45.91 & 38.41 & \textbf{50.75} & 40.21 & 82.35 & 64.33 & 89.65 & 79.50 & 23.27 & 20.41 \\ \hline
		RLGNet & \textbf{34.96} & \textbf{29.90} & \textbf{46.15} & \textbf{39.06} & 50.56 & \textbf{40.83} & \textbf{82.43} & \textbf{64.34} & \textbf{89.69} & \textbf{80.17} &  25.09& \textbf{20.81} \\ 
		\bottomrule
	\end{tabular}
	\caption{The MRR (in percentage) results of the ablation studies. Single and Multi represent single-step and multi-step reasoning, respectively. Since \textbf{Rep} can only predict repeating facts, we did not report the MRR score for \textbf{Rep}.}
	\label{tab:4}
\end{table*}
\begin{table*}[t]
	\centering
	\begin{tabular}{ c c c c c c c c }
		\toprule
		& $top_k$ & ICEWS18 & ICEWS14 & ICEWS05-15 & WIKI & YAGO  & GDELT\\ \hline
		& 5 & 34.49 & 51.94 & 42.77 & 26.69 & 84.05 & 77.67 \\
		& 10 & 40.29 & 58.26 & 47.10 & 34.26 & 90.76 & 85.37 \\
		& 20 & 44.63 & 62.90 & 49.82 & 42.35 & 92.63 & 86.96 \\
		& 30 & 46.58 & 64.80 & 50.69 & 47.05 & 92.73 & 87.03 \\
		& 100 & 49.76 & 67.59 & 51.75 & 58.41 & 92.73 & 87.04 \\
		& $\infty$ & 50.42 & 68.38 & 51.85 & 64.93 & 92.73 & 87.04 \\
		\bottomrule
	\end{tabular}
	\caption{The proportion (in percentage) of repeating facts when $top_k$ is set to different values.}
	\label{tab:5}
\end{table*}
\noindent
\subsection{Ablation Study}

To validate the impact of different modules, we conducted ablation experiments on single-step and multi-step reasoning tasks across six benchmark datasets. The results are shown in Table \ref{tab:4}. Our ablation experiments studied the impact of the Repeating History Module (\textbf{Rep}), Local History Module (\textbf{Loc}), and Global History Module (\textbf{Glo}) on performance. From the hybrid architecture perspective, the \textbf{Loc} based on the RNN architecture and the \textbf{Glo} based on the MLP architecture can complement each other in capability. For instance, \textbf{Glo}+\textbf{Loc} always outperforms the individual \textbf{Loc} or \textbf{Glo} modules. From the multi-scale perspective, information at different scales is generally beneficial to the model's performance. Analyzing from the perspective of repeating history information scale, we found that adding \textbf{Rep} to \textbf{Loc}, \textbf{Glo}, or \textbf{Loc}+\textbf{Glo} always improves the model's performance. We believe the enhancement brought by repeating history information is due to the significant proportion of repeating facts in TKG datasets, which is evident from Table \ref{tab:5}, where the proportion of repeating facts in the YAGO dataset reaches as high as 92.73\% when $top_k$ is set to $\infty$.

Regarding global and local history information scales, their performance varies due to differences between datasets (specific details can be found in Table \ref{tab:3}). Since local history information only considers facts within a unit time period, the richness of facts occurring within this period determines the performance of \textbf{Loc}. For datasets with large time spans, such as the WIKI and YAGO datasets where the unit time is one year, all facts occurring within the same year are included in one time unit. Therefore, each unit time contains more and richer facts, which is why \textbf{Loc} performs better. However, for datasets with smaller time spans, such as the GDELT dataset where the unit time is 15 minutes, \textbf{Glo} performs better because it can better grasp the characteristics of the entire dataset.

Overall, although in single-step reasoning tasks on the ICEWS05-15 and GDELT datasets, using \textbf{Glo}+\textbf{Loc}+\textbf{Rep} resulted in lower performance than \textbf{Loc}+\textbf{Rep} and \textbf{Glo}+\textbf{Rep}, the improvements in other tasks far exceeded the declines in these two sub-tasks. Therefore, we believe that the hybrid multi-scale architecture is effective in TKG reasoning tasks.

\begin{figure}[t]
	\centering
	\includegraphics[width=\linewidth]{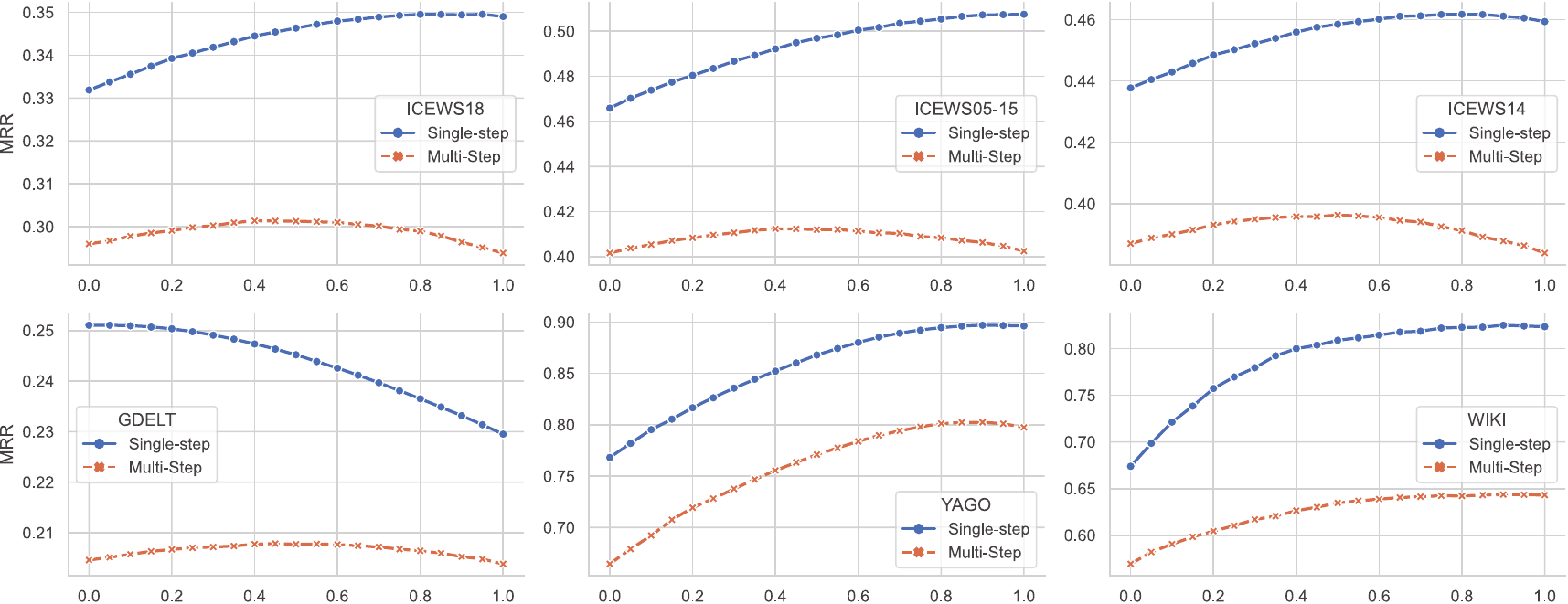}
	\caption{The impact of $\alpha$ on MRR (in percentage) results.}
	\label{fig:3}
\end{figure}

\subsection{Hyperparameter Analysis} 

We investigated the impact of the values of $\alpha$, $\omega$, and $top_k$ on the performance of RLGNet. The contributions of the Global and Local History Modules to the answers vary in single-step and multi-step reasoning tasks, which can be observed from the value of $\alpha$, as shown in Figure \ref{fig:3}. In single-step reasoning tasks, we found that on the ICEWS dataset, the performance is optimal when $\alpha$ is between 0.6 and 0.8. On the WIKI and YAGO datasets, the performance peaks when $\alpha$ is between 0.8 and 1. On the GDELT dataset, the optimal range for $\alpha$ is between 0 and 0.2. Therefore, we set $\alpha$ to 0.8 and 0.1 for ICEWS and GDELT, respectively, and 0.9 for WIKI and YAGO.

In multi-step reasoning tasks, the value of $\alpha$ usually needs to be smaller than in single-step reasoning tasks to achieve optimal performance. This indicates that the Global History Module based on MLP is more important in multi-step reasoning, whereas the Local History Module based on RNN is more crucial in single-step reasoning. The experimental results also demonstrate the unique advantages of the methods based on MLP and RNN architectures, validating our hybrid architecture design's ability to leverage the strengths of both. 

\begin{figure}[t]
	\centering
	\includegraphics[width=\linewidth]{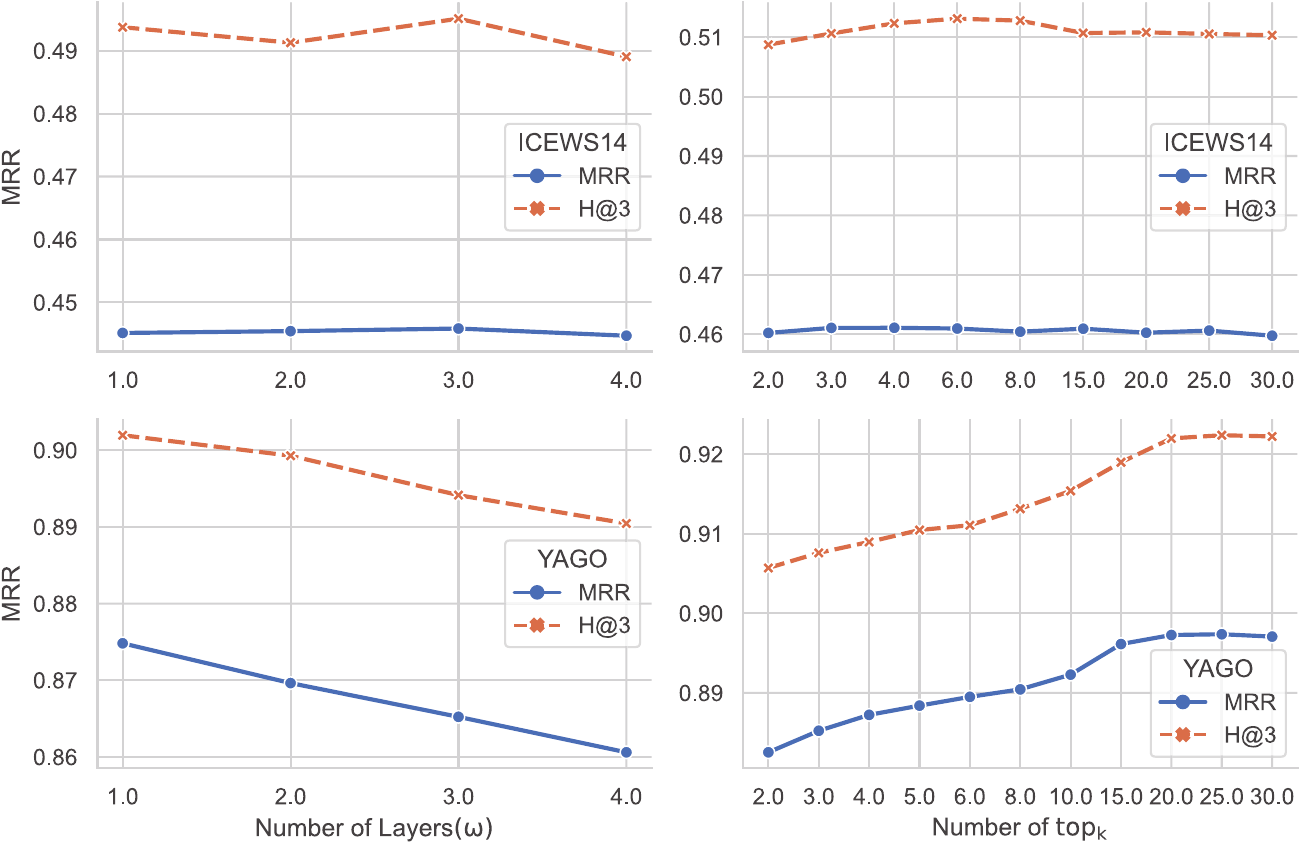}
	\caption{The impact of $\omega$ and $top_k$ on MRR(in percentage) results in ICEWS14 and YAGO.}
	\label{fig:4}
\end{figure}

Additionally, we studied the impact of $\omega$ and $top_k$ values on the performance of RLGNet on the ICEWS14 and YAGO datasets, as shown in Figure \ref{fig:4}. As $\omega$ increases, the performance of the model on ICEWS14 fluctuates slightly, while the performance on the YAGO dataset rapidly declines. Therefore, we set $\omega$ to 1 for all datasets. In the YAGO dataset, setting $top_k$ at a higher value yields better performance; however, changes in $top_k$ do not significantly affect the performance on the ICEWS14 dataset. Therefore, we set $top_k$ for all datasets to 20.

\subsection{Effectiveness of Ensemble Learning Strategy}

To demonstrate the effectiveness of the ensemble learning strategy, we compared the performance of different modules using ensemble learning and joint learning strategies on the YAGO and ICEWS14 datasets, as shown in Table \ref{tab:6}. In joint learning, all modules share the embeddings of entities and relationships, with other settings being the same as those in the ensemble learning strategy. We found that the performance of any module combination using the ensemble learning strategy was generally superior to that of the joint learning strategy. This is because, in TKG, the patterns of facts change over time, making it inevitable that there will be noisy data in the historical information. Although joint learning, through weight sharing, allows the model to perform better during the training phase, it also leads to fitting the noisy data, thereby impairing the model's ability to predict future events. Therefore, we believe that the ensemble learning strategy is an effective approach for inference tasks in TKGs.

\begin{table}[t]
	\centering
	\begin{tabular}{ c c c c c c}
		\toprule
		& \multirow{2}{*}{Model}  & \multicolumn{2}{c}{ICEWS14}  & \multicolumn{2}{c}{YAGO} \\
		\cmidrule(lr){3-4}\cmidrule(lr){5-6}
		& & Ensem & Joint & Ensem & Joint  \\ \hline
		& \textbf{Glo}+\textbf{Loc} & 45.56 & 44.98 & 87.38 & 87.38 \\
		& \textbf{Loc}+\textbf{Rep} & 45.91 & 45.10 & 89.65 & 88.95\\
		& \textbf{Glo}+\textbf{Rep} & 43.74 & 42.77 & 78.85 & 82.52\\
		& \textbf{Glo}+\textbf{Loc}+\textbf{Rep} & 46.15 & 45.40 & 89.69& 88.55 \\
		\bottomrule
	\end{tabular}
	\caption{The MRR (in percentage) results of the ablation studies. Ensem and Joint represent ensemble learning and joint learning, respectively.}
	\label{tab:6}
\end{table}

\section{Conclusion}

In this paper, we propose RLGNet for TKG reasoning. RLGNet is a model based on an ensemble learning strategy with a hybrid multi-scale architecture. By using Repeating, Local, and Global modules, RLGNet can capture historical information at different scales. To achieve complementary capabilities between modules and improve the model's performance in single-hop and multi-hop reasoning tasks, RLGNet adopts a hybrid architecture design: the Local and Global modules utilize RNN and MLP architectures, respectively, and their scores are combined through weighted summation. We also employ an ensemble learning strategy to further enhance the model's performance and generalization ability. Experiments on six benchmark datasets demonstrate that RLGNet outperforms existing models in most extrapolation tasks.

\section{Acknowledgments}
This work was supported by the National Natural Science Foundation of China (No.61866008).

\appendix

\bibliographystyle{kr}
\bibliography{reference}

\begin{thebibliography}{}

\bibitem[\protect\citeauthoryear{Bordes \bgroup et al\mbox.\egroup
  }{2013}]{TransE}
Bordes, A.; Usunier, N.; Garcia-Duran, A.; Weston, J.; and Yakhnenko, O.
\newblock 2013.
\newblock Translating embeddings for modeling multi-relational data.
\newblock volume~26.
\newblock Curran Associates, Inc.

\bibitem[\protect\citeauthoryear{Boschee \bgroup et al\mbox.\egroup
  }{2015}]{ICEWS18}
Boschee, E.; Lautenschlager, J.; O'Brien, S.; Shellman, S.; Starz, J.; and
  Ward, M.
\newblock 2015.
\newblock Icews coded event data.

\bibitem[\protect\citeauthoryear{Cho \bgroup et al\mbox.\egroup }{2014}]{GRU}
Cho, K.; van Merrienboer, B.; Çaglar G{\"u}lçehre; Bahdanau, D.; Bougares,
  F.; Schwenk, H.; and Bengio, Y.
\newblock 2014.
\newblock Learning phrase representations using rnn encoder–decoder for
  statistical machine translation.
\newblock In {\em Conference on Empirical Methods in Natural Language
  Processing}.

\bibitem[\protect\citeauthoryear{Deng, Rangwala, and Ning}{2020}]{Glean}
Deng, S.; Rangwala, H.; and Ning, Y.
\newblock 2020.
\newblock Dynamic knowledge graph based multi-event forecasting.
\newblock {\em Proceedings of the 26th ACM SIGKDD International Conference on
  Knowledge Discovery \& Data Mining}.

\bibitem[\protect\citeauthoryear{Dettmers \bgroup et al\mbox.\egroup
  }{2017}]{ConvE}
Dettmers, T.; Minervini, P.; Stenetorp, P.; and Riedel, S.
\newblock 2017.
\newblock Convolutional 2d knowledge graph embeddings.

\bibitem[\protect\citeauthoryear{García-Durán, Dumančić, and
  Niepert}{2018}]{Alberto2018}
García-Durán, A.; Dumančić, S.; and Niepert, M.
\newblock 2018.
\newblock Learning sequence encoders for temporal knowledge graph completion.
\newblock  4816--4821.
\newblock Association for Computational Linguistics.

\bibitem[\protect\citeauthoryear{Han \bgroup et al\mbox.\egroup }{2020}]{xERTE}
Han, Z.; Chen, P.; Ma, Y.; and Tresp, V.
\newblock 2020.
\newblock xerte: Explainable reasoning on temporal knowledge graphs for
  forecasting future links.
\newblock {\em ArXiv} abs/2012.15537.

\bibitem[\protect\citeauthoryear{Han \bgroup et al\mbox.\egroup
  }{2021a}]{TANGO}
Han, Z.; Ding, Z.; Ma, Y.; Gu, Y.; and Tresp, V.
\newblock 2021a.
\newblock Learning neural ordinary equations for forecasting future links on
  temporal knowledge graphs.
\newblock In {\em Conference on Empirical Methods in Natural Language
  Processing}.

\bibitem[\protect\citeauthoryear{Han \bgroup et al\mbox.\egroup }{2021b}]{TLDR}
Han, Z.; Ding, Z.; Ma, Y.; Gu, Y.; and Tresp, V.
\newblock 2021b.
\newblock Learning neural ordinary equations for forecasting future links on
  temporal knowledge graphs.
\newblock In {\em Conference on Empirical Methods in Natural Language
  Processing}.

\bibitem[\protect\citeauthoryear{Jiang \bgroup et al\mbox.\egroup
  }{2016}]{TTransE}
Jiang, T.; Liu, T.; Ge, T.; Sha, L.; Chang, B.; Li, S.; and Sui, Z.
\newblock 2016.
\newblock Towards time-aware knowledge graph completion.
\newblock In {\em International Conference on Computational Linguistics}.

\bibitem[\protect\citeauthoryear{Jin \bgroup et al\mbox.\egroup
  }{2020}]{Re-net}
Jin, W.; Qu, M.; Jin, X.; and Ren, X.
\newblock 2020.
\newblock Recurrent event network: Autoregressive structure inferenceover
  temporal knowledge graphs.
\newblock  6669--6683.
\newblock Association for Computational Linguistics.

\bibitem[\protect\citeauthoryear{Leblay and Chekol}{2018}]{WIKI}
Leblay, J., and Chekol, M.~W.
\newblock 2018.
\newblock Deriving validity time in knowledge graph.
\newblock  1771--1776.
\newblock ACM Press.

\bibitem[\protect\citeauthoryear{Li \bgroup et al\mbox.\egroup
  }{2021a}]{CluSTeR}
Li, Z.; Jin, X.; Guan, S.; Li, W.; Guo, J.; Wang, Y.; and Cheng, X.
\newblock 2021a.
\newblock Search from history and reason for future: Two-stage reasoning on
  temporal knowledge graphs.
\newblock {\em ArXiv} abs/2106.00327.

\bibitem[\protect\citeauthoryear{Li \bgroup et al\mbox.\egroup
  }{2021b}]{RE-GCN}
Li, Z.; Jin, X.; Li, W.; Guan, S.; Guo, J.; Shen, H.; Wang, Y.; and Cheng, X.
\newblock 2021b.
\newblock Temporal knowledge graph reasoning based on evolutional
  representation learning.
\newblock  408--417.
\newblock Association for Computing Machinery, Inc.

\bibitem[\protect\citeauthoryear{Li \bgroup et al\mbox.\egroup }{2022}]{CEN}
Li, Z.; Guan, S.; Jin, X.; Peng, W.; Lyu, Y.; Zhu, Y.; Bai, L.; Li, W.; Guo,
  J.; and Cheng, X.
\newblock 2022.
\newblock Complex evolutional pattern learning for temporal knowledge graph
  reasoning.
\newblock  290--296.
\newblock Association for Computational Linguistics.

\bibitem[\protect\citeauthoryear{Li, Sun, and Zhao}{2022}]{TiRGN}
Li, Y.; Sun, S.; and Zhao, J.
\newblock 2022.
\newblock Tirgn: Time-guided recurrent graph network with local-global
  historical patterns for temporal knowledge graph reasoning.
\newblock  2152--2158.

\bibitem[\protect\citeauthoryear{Liu \bgroup et al\mbox.\egroup
  }{2021}]{TLogic}
Liu, Y.; Ma, Y.; Hildebrandt, M.; Joblin, M.; and Tresp, V.
\newblock 2021.
\newblock Tlogic: Temporal logical rules for explainable link forecasting on
  temporal knowledge graphs.
\newblock In {\em AAAI Conference on Artificial Intelligence}.

\bibitem[\protect\citeauthoryear{Liu \bgroup et al\mbox.\egroup }{2023}]{RETIA}
Liu, K.; Zhao, F.; Xu, G.; Wang, X.; and Jin, H.
\newblock 2023.
\newblock Retia: Relation-entity twin-interact aggregation for temporal
  knowledge graph extrapolation.
\newblock  1761--1774.
\newblock IEEE.

\bibitem[\protect\citeauthoryear{Mahdisoltani, Biega, and
  Suchanek}{2015}]{YAGO}
Mahdisoltani, F.; Biega, J.~A.; and Suchanek, F.~M.
\newblock 2015.
\newblock Yago3: A knowledge base from multilingual wikipedias.

\bibitem[\protect\citeauthoryear{N. \bgroup et al\mbox.\egroup }{2018}]{R-GCN}
N., T.; Peter, B.; van den Berg~Rianne; Ivan, T.; Michael, W. M.~S.; and Kipf.
\newblock 2018.
\newblock Modeling relational data with graph convolutional networks.
\newblock  593--607.
\newblock Springer International Publishing.

\bibitem[\protect\citeauthoryear{Nickel, Tresp, and Kriegel}{2011}]{RESCAL}
Nickel, M.; Tresp, V.; and Kriegel, H.-P.
\newblock 2011.
\newblock A three-way model for collective learning on multi-relational data.

\bibitem[\protect\citeauthoryear{Tone}{2015}]{GDELT}
Tone, A.
\newblock 2015.
\newblock Global data on events, location and tone (gdelt).

\bibitem[\protect\citeauthoryear{Trivedi \bgroup et al\mbox.\egroup
  }{2017}]{Know-Evolve}
Trivedi, R.~S.; Dai, H.; Wang, Y.; and Song, L.
\newblock 2017.
\newblock Know-evolve: Deep temporal reasoning for dynamic knowledge graphs.
\newblock In {\em International Conference on Machine Learning}.

\bibitem[\protect\citeauthoryear{Trivedi \bgroup et al\mbox.\egroup
  }{2019}]{DyRep}
Trivedi, R.~S.; Farajtabar, M.; Biswal, P.; and Zha, H.
\newblock 2019.
\newblock Dyrep: Learning representations over dynamic graphs.
\newblock In {\em International Conference on Learning Representations}.

\bibitem[\protect\citeauthoryear{Trouillon \bgroup et al\mbox.\egroup
  }{2016}]{ComlEX}
Trouillon, T.; Welbl, J.; Riedel, S.; Gaussier, {\'E}.; and Bouchard, G.
\newblock 2016.
\newblock Complex embeddings for simple link prediction.
\newblock {\em ArXiv} abs/1606.06357.

\bibitem[\protect\citeauthoryear{Wang \bgroup et al\mbox.\egroup
  }{2014}]{TransH}
Wang, Z.; Zhang, J.; Feng, J.; and Chen, Z.
\newblock 2014.
\newblock Knowledge graph embedding by translating on hyperplanes.

\bibitem[\protect\citeauthoryear{Xu \bgroup et al\mbox.\egroup }{2023}]{CENET}
Xu, Y.; Ou, J.; Xu, H.; and Fu, L.
\newblock 2023.
\newblock Temporal knowledge graph reasoning with historical contrastive
  learning.

\bibitem[\protect\citeauthoryear{Yang \bgroup et al\mbox.\egroup
  }{2014}]{DistMult}
Yang, B.; tau Yih, W.; He, X.; Gao, J.; and Deng, L.
\newblock 2014.
\newblock Embedding entities and relations for learning and inference in
  knowledge bases.
\newblock {\em CoRR} abs/1412.6575.

\bibitem[\protect\citeauthoryear{Ye \bgroup et al\mbox.\egroup }{2019}]{VR-GCN}
Ye, R.; Li, X.; Fang, Y.; Zang, H.; and Wang, M.
\newblock 2019.
\newblock A vectorized relational graph convolutional network for
  multi-relational network alignment.
\newblock  4135--4141.
\newblock AAAI Press.

\bibitem[\protect\citeauthoryear{Zhen \bgroup et al\mbox.\egroup
  }{2018}]{Conv-TransE}
Zhen, M.; Wang, J.; Zhou, L.; Fang, T.; and Quan, L.
\newblock 2018.
\newblock End-to-end structure-aware convolutional networks for knowledge base
  completion.
\newblock {\em Proceedings of the ... AAAI Conference on Artificial
  Intelligence. AAAI Conference on Artificial Intelligence} 33:3060--3067.

\bibitem[\protect\citeauthoryear{Zhong}{2021}]{TITer}
Zhong, Y. M. Z. H. K. H. H. S.~J.
\newblock 2021.
\newblock Timetraveler: Reinforcement learning for temporal knowledge graph
  forecasting.

\bibitem[\protect\citeauthoryear{Zhu \bgroup et al\mbox.\egroup
  }{2021}]{CyGNet}
Zhu, C.; Chen, M.; Fan, C.; Cheng, G.; and Zhang, Y.
\newblock 2021.
\newblock Learning from history: Modeling temporal knowledge graphs with
  sequential copy-generation networks.
\newblock {\em Proceedings of the AAAI Conference on Artificial Intelligence}
  35:4732--4740.

\end{thebibliography}

\end{document}